\documentclass[runningheads]{llncs}
\usepackage[T1]{fontenc}
\usepackage{amsmath}
\usepackage{amssymb}
\usepackage{graphicx,verbatim}
\usepackage{url}
\usepackage[hidelinks]{hyperref}
\usepackage{orcidlink}
\makeatletter
\newcommand{\printfnsymbol}[1]{%
  \textsuperscript{\@fnsymbol{#1}}%
}
\makeatother

\begin{document}
\title{Steerable Anatomical Shape Synthesis with Implicit Neural Representations}


\author{
Bram de Wilde\inst{1,3},\thanks{These authors contributed equally to this work.}\orcidlink{0000-0003-1890-8714} \and
Max T. Rietberg\inst{2},\printfnsymbol{1}\orcidlink{0000-0001-8959-6477} \and
Guillaume Lajoinie\inst{1}\orcidlink{0000-0002-8226-7301} \and
Jelmer M. Wolterink\inst{3}\orcidlink{0000-0001-5505-475X}
}
\authorrunning{Bram de Wilde et al.}
\institute{
Physics of Fluids Group, Technical Medical (TechMed) Centre, University of Twente, Enschede, The Netherlands
\and
Multi-Modality Medical Imaging, Technical Medical (TechMed) Centre, University of Twente, Enschede, The Netherlands
\and
Department of Applied Mathematics, Technical Medical (TechMed) Centre, University of Twente, Enschede, The Netherlands\\
\email{contact@bramdewilde.com}}

\maketitle              
\begin{abstract}
Generative modeling of anatomical structures plays a crucial role in virtual imaging trials, which allow researchers to perform studies without the costs and constraints inherent to in vivo and phantom studies.
For clinical relevance, generative models should allow targeted control to simulate specific patient populations rather than relying on purely random sampling.
In this work, we propose a steerable generative model based on implicit neural representations. 
Implicit neural representations naturally support topology changes, making them well-suited for anatomical structures with varying topology, such as the thyroid.
Our model learns a disentangled latent representation, enabling fine-grained control over shape variations.
Evaluation includes reconstruction accuracy and anatomical plausibility.
Our results demonstrate that the proposed model achieves high-quality shape generation while enabling targeted anatomical modifications. 


\end{abstract}
\section{Introduction}
Evaluating the performance of novel medical imaging modalities through patient trials is expensive in terms of patient risk, time, and cost \cite{FOGEL2018156}.
Simulating \textit{in vivo} trials with tissue-mimicking phantoms is common practice, but since recreating human tissue is complex, thus experiments with high mimicking accuracy remain costly \cite{McGarry_2020}.
Virtual imaging trials, where both the patient and the imaging modality are simulated, provide a compelling alternative: they allow for inexpensive, rapid and flexible testing while closely matching patient tissue \cite{abadi_virtual_2020}.
These trials are only meaningful if the underlying anatomical model of the patients is accurate, but also covers the variations seen in the population. 


In medical imaging, factors not directly related to the clinical question can strongly influence the image and its interpretation.
Examples of such confounding factors are skin color in pulse oximetry \cite{10.1117/1.JBO.29.1.010901}, BMI in X-ray \cite{doi:10.1148/radiology.217.2.r00nv35430}, or sex in PET \cite{76baa4fc60b34bc8821e0b71a1a72241}. 
Hence, when studying the influence of patient characteristics on imaging, accurate generative models provide only a partial solution.
Ideally, the generative process allows for direct control over key anatomical characteristics, such that virtual cohorts can be generated in line with the study objective.

In recent years, implicit neural representations (INRs) have emerged as a powerful and flexible platform for shape modeling \cite{park_deepsdf_2019,vedaldi_convolutional_2020}.
Surfaces are represented as the zero-level set of their signed distance function (SDF), which is modeled with a neural network.
Unlike more traditional template-based approaches, INRs are amenable to various conditioning mechanisms and naturally allow for modeling topological changes in a population \cite{dupont_data_2022,liu_learning_2022}.
Conditioning INRs on relevant shape features introduces an additional level of control for shape synthesis, which has previously been demonstrated outside of the medical domain \cite{mu_-sdf_2021,miao_manipulating_2024}. 
Furthermore, the ability of INRs to represent shapes with varying or changing topology has been previously demonstrated in biomedical applications~\cite{wiesner2024generative}.
Topological changes in anatomy can, for instance, occur due to certain diseases like tissue adhesion, renal fusion, or osteosarcoma, or due to surgical procedures like a laryngectomy. 


In this paper, we investigate steerable anatomical shape synthesis in a population with topological variations.
As a concrete use case we opt to model the thyroid gland, which is diverse in its shape and bilateral symmetry, and not topologically consistent across the patient population.
The thyroid consists of two lobes, connected by a bridge called the \textit{isthmus}.
However, up to 33\% of all patients show agenesis of the isthmus, having two separate lobes instead of a single connected thyroid \cite{ranade2008anatomical}.
We show that INRs are capable of synthesizing anatomically feasible thyroids, and additionally condition INRs on three key anatomical features (volume, isthmus cross-sectional area and symmetry) to generate and edit thyroids in a steerable way.
Finally, we experiment with a simple correlation loss term to promote feature disentanglement.




\section{Methods}


\subsection{Model}\label{sec:model}

To model three-dimensional shapes, we use coordinate-based multilayer perceptrons (MLPs) to encode the SDF.
Similar to \cite{park_deepsdf_2019}, we use a single MLP to represent multiple shape instances by conditioning the MLP on a latent code $\vec{z}$, which is concatenated to the input coordinates.
We use MLPs with 3 hidden layers of 256 nodes and ReLU activations.
For every target shape $i$, we sample the SDF value $s \in\mathbb{R}$ for a set of coordinates $\vec{x} \in \mathbb{R}^3$:

\begin{align}
X_i = \{(\vec{x}, s) : SDF_i(\vec{x})=s\}
\end{align}

The parameters $\theta$ of the MLP $f_{\theta}$ are optimized such that the model approximates the SDF for each shape $i$ when conditioned on its latent code $\vec{z}_i$:

\begin{align}
    f_{\theta}(\vec{x}, \vec{z}_i) \approx SDF_i(\vec{x})
\end{align}

\subsection{Training}

Both the parameters $\theta$ and the latent codes $\vec{z}_i$ are optimized using the mean squared error loss on the predicted SDF values.
Additionally, we apply $L_2$ regularization to the latent codes, leading to the following total loss $\mathcal{L}$:

\begin{align}
    \mathcal{L}(f_\theta, X, \vec{z}) = \sum_{\vec{x}\in X} ||f_{\theta}(\vec{x}, \vec{z}) - SDF(\vec{x})||_2^2 + \lambda ||\vec{z}||_2^2\label{loss:default}
\end{align}

\subsection{Disentanglement}

Training a model as described in Section \ref{sec:model} does not impose any structure on the latent space. Hence, randomly sampled codes provide novel samples, but there is no easy way to control anatomical aspects, such as the volume of the generated shape.
To promote steerability of output shape characteristics, we split the latent code $\vec{z}_i$ for each shape into a fixed part $\vec{z}_{i,fixed}$, which is not updated during training, and a trainable part $\vec{z}_{i,trainable}$.
By letting $\vec{z}_{i,fixed}$ represent anatomical features, the model is directly conditioned on them, such that we can investigate disentanglement of $\vec{z}_{i,fixed}$ with respect to the trainable features.
If $\vec{z}_{i,fixed}$ is, for instance, set to the volume of each shape $i$, then $\vec{z}_{i,trainable}$ will ideally model all shape changes \textit{but} the volume.
Concretely, we investigate the two following disentanglement strategies:

\subsubsection{Fixed conditioning} The anatomical feature(s) of interest are added as fixed latent code features and training is done as described in Section \ref{sec:model}. This is a baseline approach to investigate how much the model disentangles the features by itself without any special strategies.

\subsubsection{Correlation loss} In addition to the \textbf{Fixed conditioning} strategy, we calculate a correlation loss term at the end of each epoch to update the latent codes.
The loss term promotes disentanglement of the fixed features from the trainable features.
Specifically, the loss term calculates the mean Pearson correlation coefficient between the fixed and trainable latent features:

\begin{align}
    \mathcal{L}_{corr} = \frac{1}{N}\sum_j^N \left|\frac{\text{Cov}(\vec{z}_{fixed}, \vec{z}_{trainable,j})}{\sigma_{fixed}\sigma_{trainable}}\right|\label{loss:corr}
\end{align}

Here $N$ is the number of trainable latent features, $j$ is the latent feature index.
If a model is perfectly disentangled, the fixed feature should have no correlation to any of the trainable dimensions, and hence have a low loss value.

\subsection{Inference}

Shapes are synthesized by conditioning the model on a latent code $\vec{z}$ and sampling the predicted SDF values $f_{\theta}(\vec{x},\vec{z})$ for $\vec{x} \in [0,1]^3$, where $\vec{x}$ is sampled on a $64^3$ grid.
Meshes are obtained with marching cubes.
Novel samples are synthesized by sampling randomly from the latent space.
For each trainable latent dimension, we fit and draw random values from a normal distribution.
For fixed latent dimensions in conditioned models, we sample directly from the distribution seen in the dataset, such that the synthesized population follows the training population.

\subsection{Data}

\begin{figure}[t]
    \centering
    \includegraphics[width=\linewidth]{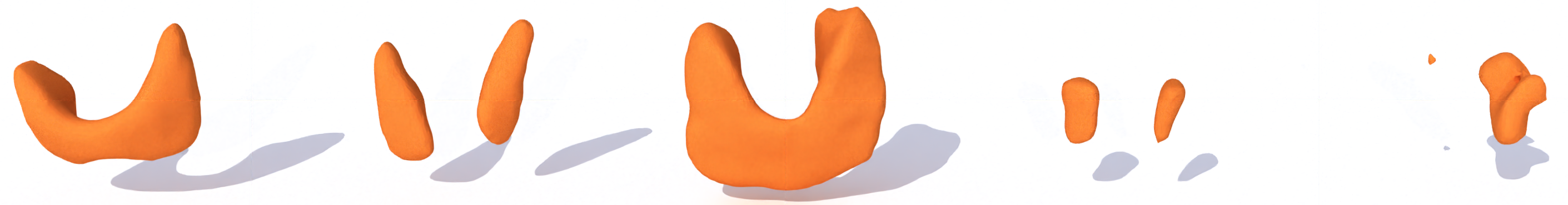}
    \caption{Examples from the dataset illustrating the variety in thyroid anatomy. From left to right: (1) a typical connected thyroid, (2) a typical split thyroid, (3) a very large thyroid, (4) a very small thyroid, (5) a highly asymmetric thyroid.}
    \label{fig:train_5_samples}
\end{figure}

We collect thyroid shapes from the TotalSegmentator training dataset \cite{wasserthal_totalsegmentator_2023}.  
The dataset consists of 1228 CT scans on which 117 structures, including the thyroid gland, have been annotated.
Of these 1228 scans, there are 415 scans where the thyroid gland is fully in the field of view and is completely annotated.
We additionally collect the trachea shape for each thyroid to center the meshes. 
Due to the diverse nature of the dataset, not all cases are annotated consistently.
After a check for annotation quality, which includes checking for a number of connected components and watertightness, we discard 62 cases.
Finally, we train all models on a set of 353 thyroids.
Representative examples from the dataset are shown in Figure~\ref{fig:train_5_samples}.

We choose to center all meshes on the trachea, instead of on thyroid center of mass, because asymmetric thyroids would give off-center results.
To allow for centering, we first convert all binary voxel grid thyroid masks to meshes with marching cubes, and then center them on the center of mass of the trachea.
The meshes are finally normalized to the largest extent in the dataset in each dimension, such that all shapes fall within a common unit cube, but size variations are preserved.

For each thyroid mesh, we sample the SDF values for 50,000 coordinates for training.
40,000 coordinates are randomly sampled on the mesh surface and hence have an SDF of 0.
The remaining 10,000 points are a random sample of the 40,000 surface points plus a randomly sampled displacement in each direction from a Gaussian distribution with a standard deviation of 0.1.
We publish the processed meshes, pre-sampled SDF values and our source code online.\footnote{\href{https://github.com/MIAGroupUT/steerable-shape-synthesis}{https://github.com/MIAGroupUT/steerable-shape-synthesis}}

\subsection{Validation}

To validate that the baseline model is able to capture the anatomical variations in the dataset, we evaluate reconstruction accuracy with the Chamfer distance between reconstructed and reference meshes.
To validate that the baseline model can also \textit{generate} meaningful novel samples, we validate that they are anatomically realistic with respect to three key anatomical features, of which we compare training to generated population. 
Specifically, the anatomical features we consider are thyroid \textbf{volume}, \textbf{isthmus area}, and \textbf{symmetry}. 

Volume is a natural measure for size variations. Furthermore, isthmus area allows for a continuous description from a split thyroid (area=0) to a connected one. 
It is calculated as the thyroid area in the midsagittal plane.
Finally, to characterize the large variation in symmetry in the dataset, we flip one half of the thyroid onto the other by mirroring at the midsagittal plane, and calculating the intersection over union between both halves.
In this measure, a perfectly symmetric thyroid has a symmetry score of 1, whereas a completely asymmetric thyroid has a score of 0.

To validate steerability of the conditioned models (\textit{fixed} and \textit{correlation}), we randomly generate 1000 thyroids and evaluate the correlation between the conditioned anatomical features and the actual generated features with the Pearson correlation coefficient (PCC).

\section{Results}

All models are trained with the Adam optimizer with a learning rate of $3\cdot10^{-4}$.
Each model is trained for 10,000 epochs, where each epoch consists of 1000 coordinate-SDF pairs for each shape, sampled randomly from $X_i$.
Models are trained with a trainable latent code of size $N=64$, with their values initialized randomly from $\mathcal{N}(0, 0.01^2)$. Throughout the results we compare three different models:

\begin{itemize}
    \item \textit{Baseline} A model conditioned only on a trainable latent code.
    \item \textit{Fixed} The baseline model, but additionally conditioned on volume, isthmus area, and symmetry. This model is trained with the default loss \eqref{loss:default}.
    \item \textit{Correlation} The fixed model, with the correlation loss \eqref{loss:corr} added.
\end{itemize}

\subsection{Reconstruction quality}
The mean Chamfer distance between training meshes and their corresponding reconstructed meshes were 1.56 $\pm$ 0.38 mm (std) for the \textit{baseline} model,  1.63 $\pm$ 0.12 mm for the \textit{fixed} model, and 1.60 $\pm$ 0.06 mm for the \textit{correlation model}.
This shows that the baseline model is able to fit all shapes in the dataset well, and that adding extra conditioning to the latent code does not impact reconstruction quality.


\subsection{Generation quality}
\begin{figure}[t]
    \centering
    \includegraphics[width=\linewidth]{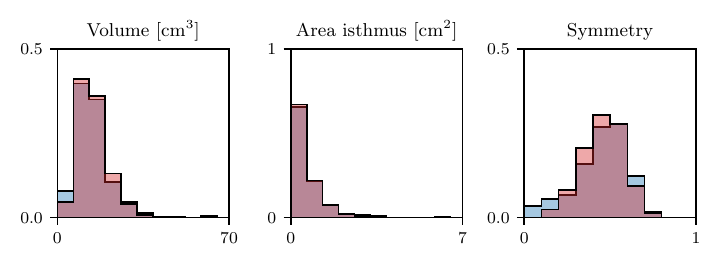}
    \caption{Comparison between volume, isthmus area and symmetry for the training data (blue) and 1000 randomly generated meshes using the \textit{baseline} model (red).}
    \label{fig:validation}
\end{figure}

To demonstrate the generative performance of the \textit{baseline} model, 1000 meshes were randomly generated.
The generated distributions of volume, isthmus area and symmetry compared to the training distributions are shown in Figure~\ref{fig:validation}.
These results indicate that our baseline approach is able to generate thyroids which cover the entire training distribution.
Moreover, the learned latent space follows the anatomical distribution of the training data.


\subsection{Steerability}

\begin{figure}[ht]
    \centering
    \includegraphics{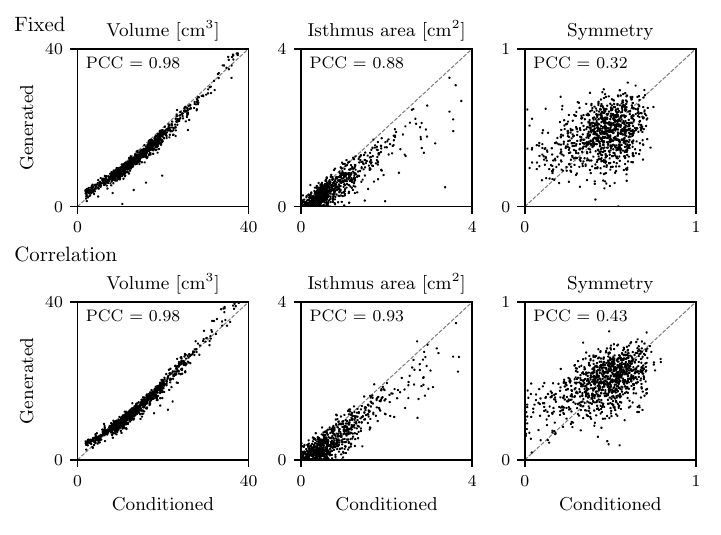}
    \caption{Correlation between conditioned and generated anatomical features for the \textit{fixed} model (top row) and the \textit{correlation} model (bottom row). The inset shows the Pearson correlation coefficient.}
    \label{fig:correlation}
\end{figure}

To evaluate to which extent latent conditioning on volume, isthmus area and symmetry works, we generate 1000 random meshes for both the \textit{fixed} and the \textit{correlation} model.
Figure~\ref{fig:correlation} shows the correlation between features the model is conditioned on and the actual generated features for the both models.
Generated volume correlates well with the conditioned volume with a PCC of 0.98 for both models, indicating that volume is a relatively easy feature to disentangle.
Isthmus area and especially symmetry appear more complex features to disentangle, and here the use of the correlation loss improves the PCC in both cases: from 0.88 to 0.93 for the isthmus area, and from 0.32 to 0.43 for symmetry.

\begin{figure}[h]
    \centering
    \includegraphics[width=\linewidth]{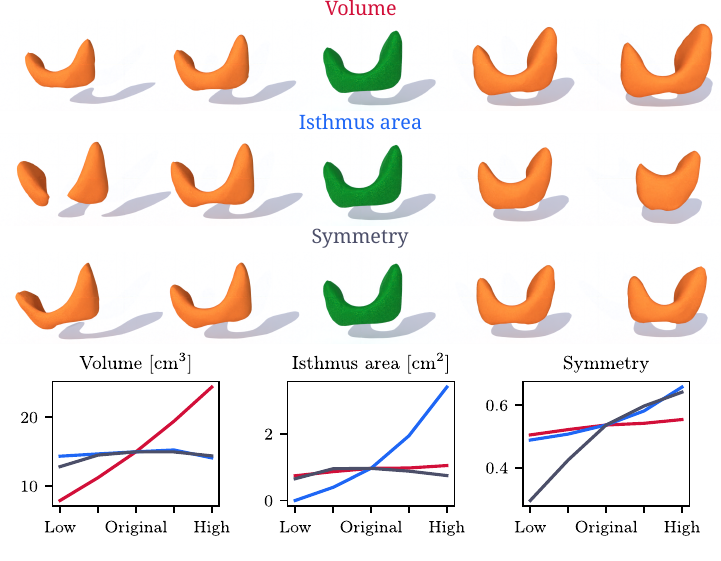}
    \caption{Editing a training mesh (middle column, green) by independently varying volume (red), isthmus area (blue) and symmetry (grey). The plots show each anatomical feature for each of the rows, demonstrating that features can be independently varied.}
    \label{fig:edit}
\end{figure}

To investigate the ability of the \textit{correlation} model to independently vary specific features, we show the editing of a particular training mesh in Figure~\ref{fig:edit}. 
While varying a specific feature, the other features show almost no difference, with exception of the isthmus area when increasing the symmetry value.
We hypothesize this might be because the cross-sectional area increases when you tilt a thyroid, i.e., make it less symmetrical.
This could be a reason why it is harder for the model to disentangle symmetry from the isthmus area than from the volume.

\section{Discussion}

We show that INRs are capable of synthesizing anatomically accurate thyroid glands, including the ability to model topological changes across a patient population.
By conditioning the models on volume, isthmus area and symmetry, it is possible to synthesize thyroids in a controlled manner.
Adding a loss to minimize correlation between fixed and trainable features gives improvement in terms of correlation between conditioned and generated features, indicating it might help the model to disentangle these features in latent space.
The resulting model can be used to generate patient cohorts with specific anatomical characteristics and allows editing anatomical features of existing thyroids with a high degree of independence.

Both Figure \ref{fig:correlation} and \ref{fig:edit} show that the three anatomical features we include are not equally complex for the model.
Volume is fitted well by both models, but isthmus area is already more difficult, which might be due to fact that it is a more local feature.
Symmetry is clearly the most complex, although Figure \ref{fig:validation} and \ref{fig:edit} show that the models can model the feature to some extent.
In order for the model to disentangle symmetry, it arguably needs to learn some representation of that feature, which might be difficult for our relatively small MLP models.

While the thyroid is a complex use-case due to variations in topology and symmetry, the shapes themselves are still relatively smooth.
Whether our approach generalizes to organs with sharper features that require a representation containing high frequencies is an interesting direction for future research.
In this case, investigating alternative model architectures such as Siren, which has been shown to have a stronger capacity for representing sharp features, is probably a fruitful direction \cite{sitzmann_implicit_2020}.
  

Another interesting avenue for future investigation is extending our approach to a multi-organ setting.
In this setting, a model could be conditioned on a unique latent code for each output organ.
The generative process can then be steered on an organ-level basis, while still providing a coherent set of generated organs, respecting the anatomical hierarchy.
Using the proposed correlation loss could help models to disentangle organ representations in latent space, keeping their representation independent, and allowing for imposing variations in single organs.

\newpage
\bibliographystyle{splncs04}
\bibliography{bib}
%




\end{document}